# Biomarker CD46 Detection in Colorectal Cancer Data based on Wavelet Feature Extraction and Genetic Algorithm

Yihui Liu[1], Uwe Aickelin[1], Jan Feyereisl[1], and Lindy G Durrant[2]

[1] School of Computer Science, University of Nottingham, UK
[2] Academic Department of Clinical Oncology, Institute of Immunology, Infections and Immunity, City Hospital, University of Nottingham, UK

**Corresponding author:**

Yihui Liu

Institute of Intelligent Information Processing

Shandong Polytechnic University,

Jinan, China, 250013

 Email: yihui_liu_2005@yahoo.co.uk;

          yxl@spu.edu.cn

Tel: +86 (0) 53189631256

Abstract:

Biomarkers which predict patient's survival can play an important role in medical diagnosis and treatment. How to select the significant biomarkers from hundreds of protein markers is a key step in survival analysis. In this paper a novel method is proposed to detect the prognostic biomarkers of survival in colorectal cancer patients using wavelet analysis, genetic algorithm, and Bayes classifier. One dimensional discrete wavelet transform (DWT) is normally used to reduce the dimensionality of biomedical data. In this study one dimensional continuous wavelet transform (CWT) was proposed to extract the features of colorectal cancer data. One dimensional CWT has no ability to reduce dimensionality of data, but captures the missing features of DWT, and is complementary part of DWT. Genetic algorithm was performed on extracted wavelet coefficients to select the optimized features, using Bayes classifier to build its fitness function. The corresponding protein markers were located based on the position of optimized features. Kaplan-Meier curve and Cox regression model were used to evaluate the performance of selected biomarkers. Experiments were conducted on colorectal cancer dataset and several significant biomarkers were detected. A new protein biomarker CD46 was found to significantly associate with survival time.

# 1. Introduction

Survival analysis involves the estimation of the distribution of time it takes for death to occur depending on the biology of the disease. It allows clinicians to plan a suitable treatment and counsel patients about their prognosis. In medical domains, survival analysis is mainly based on Kaplan-Meier (KM) estimator and Cox proportional hazards regression model [1,2], which are used to evaluate the performance of prognostic markers. However how to rank these biomarkers, is a key step in survival analysis. Normally, the selection of biomarkers is based on medical knowledge and the diagnosis of the clinician [1, 2]. This may ignore potential biomarkers. Machine learning algorithms have been widely used in biomarker analysis of high dimensional medical data, such as microarray data [3,4,5] or mass spectrometry data [6,7]. Despite the potential advantages over standard statistical methods, their applications to survival analysis are rare due to the difficulty in dealing with censored data [8]. Recent research has shown that machine learning methods, such as neural network [9,10], Bayesian network [11], decision tree and Naïve Bayes classifier [8], are used to improve the survival model. However, none of these methods deals with the biomarker selection in survival analysis.

In this study we propose a novel method of biomarker selection based on one dimensional continuous wavelet transform (CWT). Normally one dimensional discrete wavelet transform (DWT) is used to reduce dimensionality in the analysis of high dimensional biomedical data [12,13]. In biomarker detection, the feature space must have the corresponding relationship with original data space to locate the detected biomarker based on detected features. One dimensional CWT detects the feature of data at every scale and position, and keeps local property of the original data. Wavelet feature vector of CWT has the same length as the original data, and can be used to locate the biomarker in original data space.

First we perform one dimensional continuous wavelet transform at different scales on colorectal cancer data to extract the discriminant features. Then we use genetic algorithm (GA) and Bayes classifier to select the optimized features from extracted wavelet coefficients. Due to the wavelet well-known property, which reveals the local features of data (or time feature) and does not lose the position information of original data, the corresponding protein markers in the original data space are obtained based on the position of optimized wavelet features. Finally Kaplan-Meier (KM) estimator and Cox regression model were used to evaluate the performance of selected protein markers. A new protein biomarker CD46 was found to have independent prognostic significance. Recent research suggests that "the immune system might be involved in the development and progression of colorectal cancer" [1,14]. The detection of CD46 supports their deduction or conclusions.

The rest of paper is organized as follows. In section 2, we introduce the colorectal cancer data, and our proposed method is in section 3. Wavelet feature extraction for colorectal cancer data is described in section 4. In section 5, GA based on Bayes classifier is used to select the optimized features. Survival models are used to evaluate the selected biomarkers in section 6. The experiments are conducted in section 7, followed by discussion and concluding comments in section 8.

## 2. Colorectal cancer data

We use the same dataset, which Professor Lindy Durrant used in their research. It is described in Lindy Durrant's research [1, 2]. The study population cohort comprised a consecutive series of 462 archived specimens of primary invasive cases of colorectal cancer (CRC) tissue obtained from patients undergoing elective surgical resection of a histologically proven primary CRC at Nottingham University Hospitals, Nottingham, UK. The samples were collected between January 1994 and December 2000 from the established institutional tumour bank and were identified from the hospital archives. No cases were excluded unless the relevant clinicopathological material/data were unavailable. The mean follow-up period was 42 months (range 1-116) to ensure a sufficient duration of follow-up to allow meaningful assessment of the prognostic value of the markers examined. Follow-up was calculated from the date of resection of the primary tumour, and all surviving cases were censored for data analysis in December 2003. A tissue microarray of 462 colorectal tumours was stained by immunohistochemistry for markers which predict immunosurveillance/editing. There are totally 210 features.

The data has 462 samples with 210 attributions and is 462x210 data matrix. We use a simple way to do the pre-processing of our data: First, we remove those 70 features for which most patients have missing values. Second, we remove those patients, which miss any of the remaining 140 attributions. After that, we obtain a (complete) 153x140 matrix. Eighteen patients died for other causes, not related to their colorectal cancer and they were excluded from the analysis. Among the remaining135 patients, 76 patients were dead with survival time ranging from 0 to 65 months, and 59 patients were alive with survival time ranging from 38 to 111 months.

The aim of the research was to find the significant biomarkers in survival analysis. Two groups of patients were identified to perform the analyses. Patients who died with survival time of less than 30 months and patients who were alive with survival time of more than 70 months. For the first group, there were 59 dead patients; for the second group, there were 31 alive patients. Among 140 attributions, only 115 of them are protein markers, others are the description of patients and medical diagnosis, such as age, survival time, TNM (Tumor, Node, Metastasis) stage and Duke stage. For this research, only protein markers were of interest in survival analysis.

Finally, there were 59x115 and 31x115 data matrix groups. Figure 1 shows two groups of data used for biomarker selection. Because the value of protein markers has a different scale, preprocessing by normalizing each protein marker and then each sample vector was done.

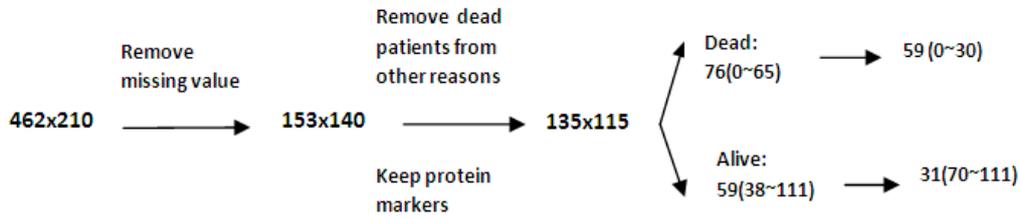

Fig. 1. Two groups of data used to select the significant protein markers. There are 59 dead patients with survival time of less than 30 months, and 31 alive patients with survival time of more than 70 months.

## 3. The proposed method

Figure 2 shows the selection process of significant biomarkers in survival analysis. First the data was transformed into wavelet space at different scales to find the most discriminant features between the two groups. Genetic algorithm was used to select the best features from extracted wavelet features and then the significant protein markers were detected based on the optimized features in wavelet space. Finally Kaplan-Meier curve and Cox regression model were performed to evaluate the performance of selected significant biomarkers.

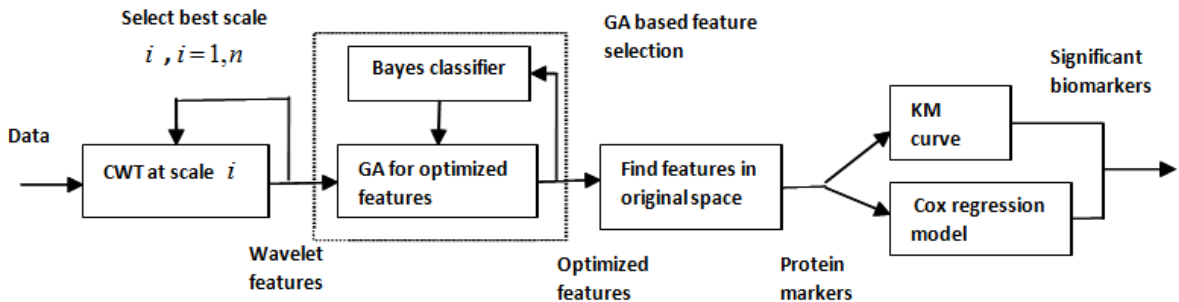

Fig. 2. The selection of significant biomarkers in survival analysis.

Normally we have feature extraction and feature selection methods for data analysis. Feature extraction is that the data is transformed into a new data space using a set of new basis, which reflects the hidden properties of data in original data space, such as principal component analysis (PCA), linear discriminant analysis (LDA), independent component analysis (ICA), and wavelet transform, etc. For PCA, the data is transformed into Eigen space, which holds the main components of data. Linear discriminant analysis extracts discriminant information by maximizing between-class variations and minimizing within-class variations. Instead of transforming uncorrelated components, like PCA and LDA, ICA attempts to achieve statistically independent components in the transform for feature extraction. For wavelet transform, the data is transformed into wavelet space, in which a set of wavelet vectors hold the different frequency properties of data. The famous property of wavelet analysis is to keep the local properties of data. Feature extraction methods of PCA, LDA, and ICA have lost the local features of data. In this research, we need select the best discriminant biomarkers, and there is no corresponding relationship between the feature space of PCA, LDA, and ICA and the biomarkers of original data. So other feature extraction methods such as PCA, LDA, and ICA are not suitable for this research. In our research, one dimensional continuous wavelet transform is used to extract the features. We first decompose the data into wavelet space of different frequencies. Then GA feature selection is used to select the best discriminant biomarkers based on wavelet features of different frequencies.

In the paper [15,16], discrete wavelet transform is used to extract the features of microarray data and protein mass spectrometry data. Microarray data and mass spectrometry data are normally of very high dimensionality, and each sample includes more than ten thousand dimensions. Wavelet coefficients of DWT are extracted to reduce the dimensionality of data, and remove redundant information. In the paper [4], in order to find the significant biomarkers, wavelet detail is reconstructed to reflect the information of data in original data space, using wavelet coefficients of DWT.

In this research, CWT is used to extract the features of colorectal cancer data. Discrete wavelet transform is non-redundant representation, which is associated with orthonormal basis. Continuous wavelet transform is redundant representation, which uses much more scale and position values than the orthonormal basis of DWT. Discrete wavelet transform has the property of dimensionality reduction, whereas CWT has no such feature and holds the same dimensionality as the original data. One dimensional CWT can be useful to detect the features of colorectal cancer data, instead DWT using discrete time-scale representation. From the view of feature extraction, CWT is complementary part of DWT, and can capture the missing features of DWT.

Feature selection methods normally have two main streams: open-loop methods and closed-loop methods [17]. The open-loop methods (filter method) select features based on between-class separability criterion, which is not involved in classification performance in the process of feature selection. The closed-loop methods, called the wrapper methods [18], select features using classification performance as a criterion of feature subset selection. Feature selection methods such as Student's t-test, Wilcoxon test, rough set [19,20], and mutual information [21,22] are open-loop methods. Normally the optimal performance means the minimal classification error. In feature selection, it has been noticed that the combinations of individual good features based on different statistic rules do not necessarily lead to good classification performance. In other words, "the m best features are not the best m features" [23]. Wrapper methods have two categories based on search strategy: greedy and randomized/stochastic [24]. Sequential backward selection (SBS) and Sequential forward selection (SFS) are the two most commonly used wrapper methods, using a greedy hill-climbing search strategy. In SBS, once a feature is removed, it is removed permanently; In SFS, once a feature is added, it is never removed. This is based on the assumption that prediction accuracy never decreases as the number of features increases. The assumption is not reliable because of search space dimensionality and overfitting. Both SBS and SFS can easily be trapped into local minima [24]. Stochastic wrapper methods, such as genetic algorithm and simulated annealing (SA), are at the forefront of research in feature subset selection. Genetic algorithm is widely used in biomarker selection [25,3,4]. In this research, genetic algorithm based on Bayes classifier is employed to select the optimized subsets of features in order to achieve the best classification performance.

## 4. Wavelet feature extraction

A wavelet is a "small wave", which has its energy concentrated in time. A wavelet system is generated from a single scaling function or wavelet by simple scaling and translation. Wavelets have a more accurate local description and separation of signal characteristics, and give a tool for the analysis of transient or time-varying signal [26]. Wavelets are widely used for image processing and feature extraction of data [27,28]. One dimensional continuous wavelet transform of signal or data $s$ is the family of wavelet coefficients $C(a,b)$, which depend on two variables of position $a$ and scale $b$. It is defined as follows:

$$C(a,b) = \int_R s(t) \frac{1}{\sqrt{a}} w(\frac{t-a}{b}) \ \mathrm{d}\,t \qquad\qquad (1)$$

where $w(t)$ is the basic wavelet and $a,b \in R$ are real continuous variables, which represent position and scale parameters of signal. Wavelet coefficients $C(a,b)$ are "resemblance degree" between the data and the wavelet at position $a$ and at scale $b$. If the coefficient value is large, the similarity is strong, otherwise it is weak.

One dimensional continuous wavelet transform is a representation of signal using much more scale and position values than an orthonormal family. Discrete wavelet transform has a discrete time-scale representation which is associated with an orthogonal basis. For DWT, approximation vectors are orthogonal to detail vectors, and detail vectors at different scales are orthogonal to each other.

Discrete wavelet transform is defined as follows:

$$C(a,b) = \int_R s(t) \frac{1}{\sqrt{a}} w(\frac{t-a}{b}) \ \mathrm{d}\,t \qquad (2)$$

$$a = k2^j, b = 2^j, (j,k) \in Z^2$$

where variable $Z$ is a set of all integers; $j$ is scale or level parameter; $k$ is the time or space location.

In this research, Daubechies wavelet db7 [29] was used for wavelet decomposition of colorectal cancer data.

The colorectal cancer data was decomposed based on one dimensional DWT. Figure 3 shows the wavelet detail coefficients and reconstructed detail at three levels for colorectal cancer data. Wavelet detail coefficients were a set of orthogonal basis, which characterize the features of colorectal cancer data at different scales. After decomposition, the dimensions of detail coefficients are 64, 38, and 25 at first three levels. The wavelet 25 detail coefficients at third level were observed to lose too much information compared to the original data space. Reconstructed detail at third level based on 25 detail coefficients may not accurately reflect the features of original data. One important property of DWT is to reduce dimensionality of data. Discrete wavelet analysis is suitable for feature extraction of very high dimensional data, such as microarray data, mass spectrometry data. But in this research, colorectal cancer data only contain 115 dimensions, and its dimensionality is not very high. Our main aim was to find the discriminant features between two groups of samples. Thus CWT was used in this research to capture the missing features of DWT.

Continuous wavelet transform extracts the features at every scale and position of data. Discrete wavelet transform at level 1, 2, and 3 equals to continuous wavelet transform at scale 2, 4, and 8 ($2^1, 2^2, 2^3$). The features of DWT at third level reflect one of CWT at scale 8. This is the reason that detail coefficients of DWT at third level lose too much information compared to the original data space.

Continuous wavelet transform at scale 1 to 5 was performed to extract the features of colorectal cancer data. Figure 4 shows the features of CWT at scale 1, 3, and 5 for colorectal cancer data. Normally noise exits at first scales of CWT. When the decomposition scale increases, less noise remains in the data. Figure 5 shows the wavelet features of DWT at level 1, 2, and 3 for the two groups of patients. Figure 6 shows the wavelet features of CWT at scale 1, 3, and 5 for the two groups of patients. For clarity, only the first 12 protein markers are shown. From Figure 5 and Figure 6, we notice that the coefficients of CWT at scale 3 show the best discriminant features between the two groups of patients. The experimental results in section 7.1 also prove that the coefficients of CWT at scale 3 achieve the best classification performance.

The shape of wavelet changes in scales, and reveals the features of signal at different derivatives. Wavelet transform can be used to calculate approximation derivatives [30,31,32]. Wavelet derivatives enhance the signal-to-noise ratio at higher order derivative calculation and retain all major properties of the conventional methods. Wavelet derivatives are much better than numerical derivatives, particularly for signals or data with lower signal-to-noise ratio (SNR). This is because SNR decreases exponentially with the increase of the derivative order in numerical derivatives. Wavelet derivative SNR does not change too much with the increase of derivative order because wavelet transform is of multiscale character, which can capture deterministic features and denoise signals or data simultaneously [31]. So the resolution of signals or data can be enhanced progressively when wavelet derivatives are used instead of numerical derivatives, and the procedure for derivative is simplified accordingly since smoothing technique is not required.

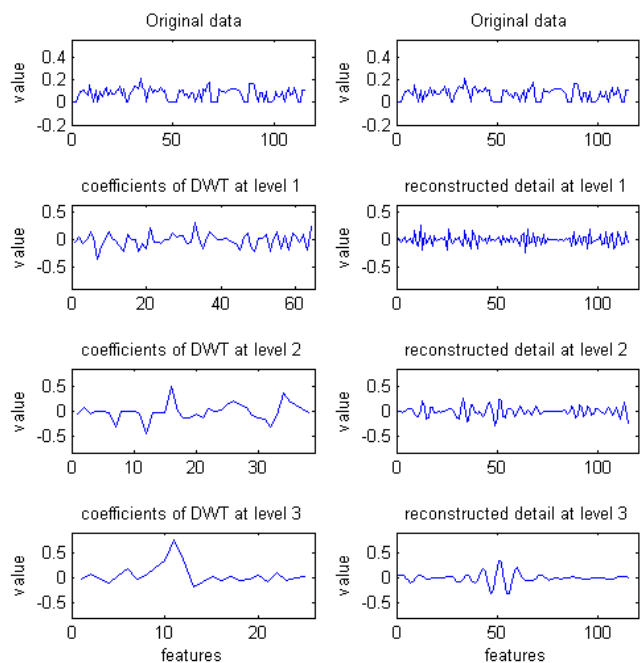

Fig. 3. Wavelet detail coefficients and reconstructed detail at three levels for colorectal cancer data. This analysis is based on DWT.

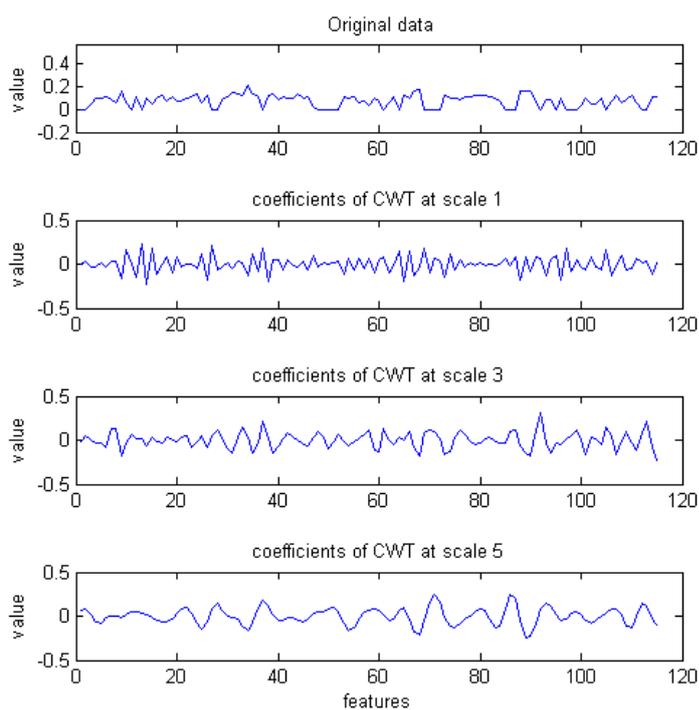

Fig. 4. Wavelet coefficients at scale 1, 3, and 5 for colorectal cancer data. This analysis is based on CWT.

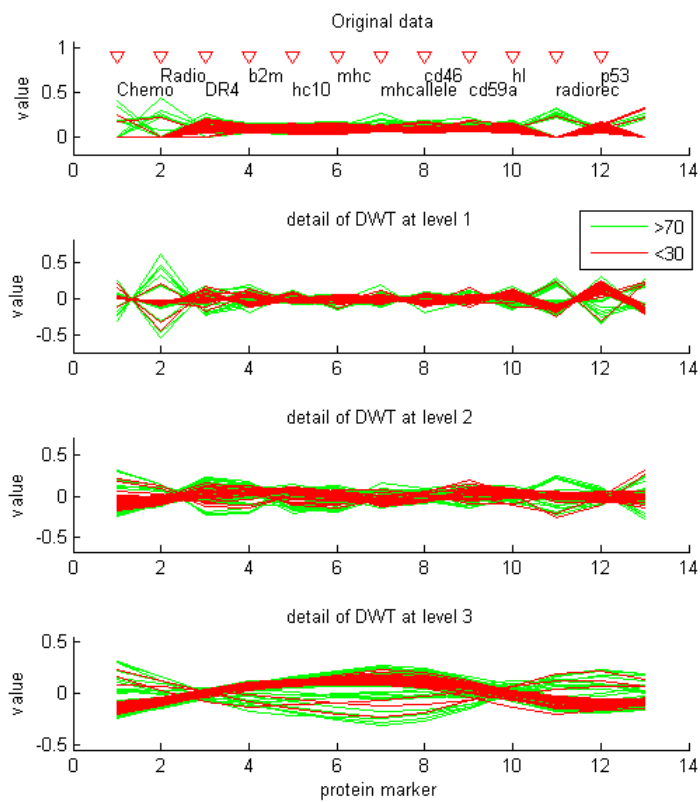

Fig. 5. The wavelet features of DWT at three levels for the two groups of patients. In order to see clearly, only first 12 features, and their corresponding protein markers are shown in the figure. The red line represents the samples for the patients who died with survival time of less than 30 months. The green line represents the samples for the patients who were alive with survival time of more than 70 months.

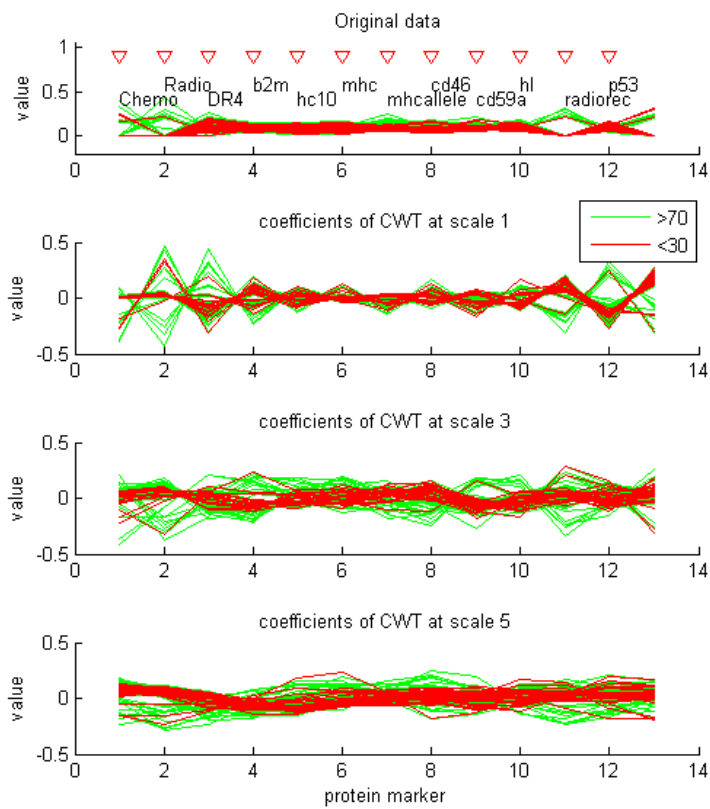

Fig. 6. The wavelet features of CWT at scale 1, 3, and 5 for the two groups of patients. Coefficients at scale 3 show the clear difference between the two groups. The red line represents the samples for the patients who died with survival time of less than 30 months. The green line represents the samples for the patients who were alive with survival time of more than 70 months.

## 5. Biomarker selection based on genetic algorithm and Bayes classifier

After wavelet feature extraction, genetic algorithm is employed to select the best features. Floating point encoding or real encoding is used in this study. Student's t-test is performed on wavelet coefficients of CWT at scale 3 to select the initialization chromosome. Uniform crossover and Gaussian mutation are performed to create next generations. In the fitness function, Bayes classifier is used to evaluate the performance of subset features, using a linear combination of the empirical error of the Bayes classifier and the a-posteriori probability.

Support vector machine and Bayes classifier are widely used in pattern recognition. In this research, Bayes classifier is used to create the fitness function. On the one hand, Bayes classifier implements quickly and simply, and is suitable for large space search using GA method. On the other hand, a-posteriori probability based on Bayes classifier is involved the computation of fitness function. When GA is running to select the best features, the whole dataset as training dataset and test dataset is used in Bayes classifier, in order to reduce the computation load.

After running GA feature selection, the best feature subsets are input into Bayes classifier to evaluate the performance, using K fold cross validation experiments. Because the famous property of wavelet feature extraction that is to keep the local property of data, biomarkers in original data space have the same index as wavelet features in wavelet space. So the corresponding significant protein markers in original data space can be located using the same index of the optimized features in wavelet space.

5.1 Bayes classifier

A Bayes classifier is a simple probabilistic classifier based on applying Bayes' theorem [33].

Given $x \in R^l$ and a set of $c$ classes, $\omega_i; i = 1,2,...,c$, the Bayes theory indicates that

$$P(\omega_i/x)p(x) = p(x/\omega_i)P(\omega_i) \qquad (3)$$

where $\quad p(x) = \sum_{i=1}^{c} p(x/\omega_i)P(\omega_i)$

where $P(\omega_i)$ is a priori probability of class $\omega_i$; $P(\omega_i/x)$ is posteriori probability of class $\omega_i$ given the vector of $x$; $p(x)$ is probability density function (pdf) of $x$; and $p(x/\omega_i)$ is the class conditional pdf of $x$ given $\omega_i$.

Let $x = [x_1, x_2,...,x_l] \in R^l$ be the features vector, whose class label is unknown. According to the Bayes decision theory, $x$ is assigned to the class $\omega_i$ if

$$P(\omega_i/x) > P(\omega_j/x), \forall j \neq i$$

or,

$$p(x/\omega_i)P(\omega_i) > p(x/\omega_j)P(\omega_j), \forall j \neq i$$

One of the most common probability density functions in practice is the Gaussian or normal density function. In this study we assume that the class conditional pdf of $x$ given $\omega_i$ follows the general multivariate normal density.

$$p(x/\omega_i) = \frac{1}{(2\pi)^{1/2}|\Sigma_i|^{1/2}} \exp\left(-\frac{1}{2}(x - \mu_i)^T \Sigma_i^{-1}(x - \mu_i)\right) \qquad (4)$$

where $\mu_i = \mathrm{E}[x]$ is the mean value of the $\omega_i$ class and $\Sigma_i$ the $l \times l$ covariance matrix is defined as

$$\Sigma_i = \mathrm{E}\left[(x - \mu_i)(x - \mu_i)^T\right]$$

$|\Sigma_i|$ denotes the determinant of $\Sigma_i$. In this research the multivariate normal density of each class is fitted with a pooled estimate of covariance.

5.2 Genetic algorithm

5.2.1 Encoding

Genetic algorithm is an evolutionary computing technique that can be used to solve problems with a vast solution space [34]. For this optimization problem, it is more natural to represent the coding variables directly as real numbers [35,36,37,38]. This means there are no differences between the coding variables and real search space. Conventionally, binary strings are used to represent the decision variables of the optimization problem in the genetic population, irrespective of the nature of the decision variables. A major drawback of binary-coded GAs is that they face difficulties when applied to problems having large search space and seeking high precision. To overcome these difficulties related to binary encoding, real encoding or floating point representation of chromosomes is used. Genetic algorithms which make use of the real encoding of chromosomes are termed as Real Coded GAs.

The use of the real encoding in the GA representation has a number of advantages over binary coding. There is no need to convert the solution variables to the binary type, and less memory is required, so the efficiency of the GA is increased. There is no loss in precision, and there is greater freedom to use different genetic operators. With floating point representation, the evaluation procedure remains the same as that in binary-coded GA, but crossover and mutation operation is done variable by variable.

5.2.2 Solution space

Genetic algorithms are used to find the best subset features to separate two classes. The feasible solution vector of this problem can be represented as $x = (x_1, x_2, \ldots x_l)$, each component or gene of which is mutually unequal and is a positive integer. Each gene is an index in the original feature set. The solution space is the subset that satisfies above constraint conditions.

We define a chromosome $x$ as a vector consisting of $l$ variables as follows.

$$x = \left\{ (x_1, \ldots, x_i, \ldots, x_j, \ldots, x_l) \middle| 1 \le \forall i \le l : 1 \le x_i \le d_{\max}; \ 1 \le i, j \le l, i \ne j : x_i \ne x_j \right\}$$

where $d_{\max}$ is the dimension number of wavelet features, i.e. the dimensionality of the original data. Because the gene is integer vector, we covert the encoding variable into one of integer formats when fitness function is calculated.

In the experiments, in order to evaluate the performance of the subset features, the chromosome length or the number of subset features $l$ is set from 1 to 20, respectively. The population size is set as follows.

$$N = \frac{d_{\max}}{l}$$

where $N$ is population number.

Student's t-test is used to select the best $l$ features to build the initial population.

5.2.3 Selection

The selection function chooses parents for the next generation, using roulette wheel and uniform sampling based on the expectation of each parent. A roulette wheel with a slot for each parent is created based on the probability of the parents, using their scaled values from the fitness function. The algorithm moves along the wheel in steps of equal size. At each step, the algorithm allocates a parent from the slot it lands on. This mechanism is fast and accurate.

5.2.4 Crossover

Uniform crossover is used in this study. A random binary vector is created. The genes where the vector is a 1 are selected from the first parent. The genes where the vector is a 0 are selected from the second parent. The genes from two parents are combined to form the child. The individuals with the best fitness values in the current generation are guaranteed to survive to the next generation. These individuals are called *elite children*. The default value of elite count is 2. The fraction of individuals in the next generation that are created by crossover, are called crossover fraction. The default value of crossover fraction is set to 0.8.

The numbers of crossover and mutation children and are calculated as follows.

$$N_c = R_c(N - N_e)$$

$$N_m = N - N_e - N_c$$

where $N$, $N_c$, $N_m$, $N_e$, and $R_c$ are the numbers of population size, crossover children, mutation children, elite children, and crossover fraction.

5.2.5 Mutation

Mutation provides genetic diversity and enables the genetic algorithm to search a wider space. Gaussian mutation is used in this study, which adds a random number created by a Gaussian distribution with mean 0 to each gene of the parent vector. The standard deviation of this distribution is determined by the parameters of scale $P_c$ and shrink $P_s$. Scale parameter $P_c$ controls the gene's search range. Shrink parameter $P_s$ controls how fast the scale $P_c$ is reduced as generations go by. The mutation child can be calculated as follows.

$$x'_k = x_k + G(0, \sigma_k) \tag{5}$$

$$\sigma_k = \sigma_{k-1}(1 - P_s \frac{k}{N_G}), \ k = 2, ..., N_G$$

$$\sigma_1 = P_c \cdot (x^u - x^l)$$

where $x'_k$ is the muted vector of $x_k$. Variable $\sigma_k$ is the standard deviation of Gaussian distribution at $k$ generation. $N_G$ and $k$ are maximum generation and the current generation number. $x^u$ and $x^l$ are upper and lower bounds of search space of $x$. The default values of both parameters of scale $P_c$ and shrink $P_s$ are 1.

5.2.6 Fitness function

We perform a Bayes classifier to design the fitness function to evaluate how well the data gets classified. A linear combination of the empirical error of the Bayes classifier or the misclassification error rate and the a-posteriori probability is employed to estimate the quality of the feature subset under examination.

The fitness function is defined as follows.

$$f(x) = e_c + e_p \tag{6}$$

where $e_c$ is the empirical classification error rate based on Bayes classifier, using the whole dataset as test samples and training samples, in order to save the computation time.

Variable $e_p$ is the defined as follows.

$$e_p = 1 - \frac{1}{n_{sam}} \left\{ \sum_{i=1}^{n_{sam}} \max \left[ P(\omega_1 \mid x_i), ..., P(\omega_c \mid x_i) \right] \right\} \tag{7}$$

where $n_{sam}$ represents the number of training sample vectors and $P(\omega_j/x_i)$ is the posteriori probability of vector $x_i$ belonging to class $\omega_j$. Assume that two subsets of $s_1$ and $s_2$ obtain the same empirical classification error rate, and the different posteriori probability of $p_1$ and $p_2$. When the condition $p_1 > p_2$ exits, subset $s_1$ is getting lower fitness value than subset $s_2$. It is clear that subset $s_1$ is better individual than subset $s_2$.

## 6. Survival analysis

### 6.1 Kaplan–Meier estimator

Kaplan–Meier (KM) analysis is a non-parametric technique for estimating time-related events, especially when not all subjects continue in the study [39]. It analyses the distribution of patient survival times following the enrolment into a study, including the proportion of alive patients up to a given time following enrolment, i.e. "censored data". "Censored data" means that the survival time for the subjects cannot be accurately determined as these patients are still alive at the time of data collection. In KM curve, a plot of the proportion of patients surviving against time has a characteristic decline (often exponential). The steepness of the curve indicates the efficacy of the treatment. The more shallow the survival curve, the more effective the treatment [40]. Log-rank test is normally used to test the statistical significance of differences between the survival curves of two different groups.

The KM estimator is nonparametric estimator of survivor function $S(t)$.

$$\hat{S}(t) = \coprod_{t_i \leq t} \left( 1 - \frac{d_i}{n_i} \right) \qquad (8)$$

where $t_i$ is the duration of study at time point $i$, $d_i$ is the number of deaths up to point $i$ and $n_i$ is number of individuals at risk just prior to $t_i$. $S$ is the product of these conditional probabilities $p_i$.

### 6.2 Cox regression model

Cox regression model is the most widely used method of survival analysis, which is not based on any assumptions concerning the nature or shape of the survival distribution [41]. Survival modelling examines the relationship between survival and one or more predictors, called covariates in survival analysis. The examination commonly has the specification of a linear-like model for the log hazard.

The Cox regression model is given by

$$h_i = \exp(\beta_1 x_{1i} + \beta_2 x_{2i} + ... + \beta_n x_{ni}) h_0(t) \qquad (9)$$

where $h_0(t)$ is called baseline hazard function, $\beta_i$ are coefficients to be solved, and $\beta_1 x_{1i} + \beta_2 x_{2i} + ... + \beta_n x_{ni}$ is the risk score or linear predictor. $x_{1i}, x_{2i}, ..., x_{ni}$ are the values of predictor variables at time $t$ for the $i^{th}$ observation.

The hazard ratio between different observations is constant and independent of time,

$$\frac{h_i(t)}{h_j(t)} = \frac{\exp(\beta' x_i)}{\exp(\beta' x_j)}$$

where $\beta = (\beta_1, \beta_2, ..., \beta_n)$ is the coefficient vector and $\beta'$ is the transpose of $\beta$. Consequently the Cox regression model is a proportional-hazards model.

# 7. Experiments and results

In this section, several experiments are conducted. In section 7.1, we compare extracted features using CWT with ones using DWT. The experimental results show that our proposed CWT method has the ability to catch the information that DWT is missing. In section 7.2, the performance of CWT features is compared with one of original data, because other feature extraction methods, such as PCA, LDA, etc., are not applicable in biomarker detection. In section 7.3, several subsets of biomarkers are selected by performing GA feature selection on CWT features at scale 3. In section 7.4, survival models of KM curve and Cox regression are used to evaluate the performance of selected biomarkers, and how the censored data affect the survival models is analysed.

### 7.1 The comparison of CWT features and DWT features

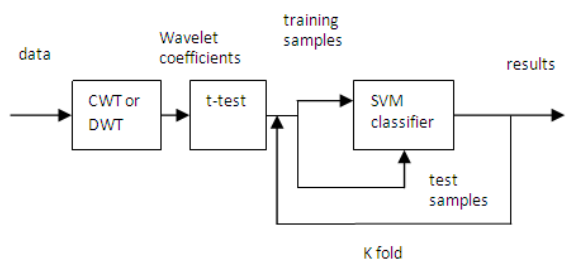

Fig. 7. The process of testing the performance of CWT features and DWT features.

In this experiment, the performance of the features of CWT and DWT were compared for the two groups of patients. They were 31 alive patients and 59 patients dead from colorectal cancer. After wavelet features of CWT and DWT were extracted, the significant features to distinguish the two groups of patients were selected based on Student's t-test [42]. The feature number was set from 1 to 20 respectively to obtain the best performance. The selected best features were input to SVM [43] classifier to evaluate the performance of classification based on 10 fold cross validation experiments running 50 times. The experimental process is shown in Figure 7.

Table 1 shows the performance of wavelet features based on CWT and DWT. For wavelet features of DWT, the best performance of 73.2% was obtained at second level based on 6 features. The 10 features of CWT at scale 3 achieved the best performance of 78.5%, and were better than the features of DWT at second level. This is because CWT was able to detect the features of colorectal cancer data at every scale and position, and does not lose information of the discriminant features between the two groups. This is especially important for the data, which was not of very high dimensionality.

Table 1. The results of wavelet features of CWT and DWT

| CWT (10 features) | DWT (6 features) |
|---|---|

| scale | Accuracy (10 fold) | Level | Accuracy (10 fold) |
|---|---|---|---|
| 1 | 0.681 | 1 (CWT=2) | 0.675 |
| 2 | 0.686 | 2 (CWT=4) | 0.732 |
| 3 | 0.785 | 3 (CWT=8) | 0.667 |
| 4 | 0.725 | 4 (CWT=16) | 0.581 |
| 5 | 0.677 | 5 (CWY=32) | 0.651 |

7.2 The comparison of CWT features and the original data

Student's t-test and Wilcoxon test [44] were used to compare the performance of CWT wavelet features with one of the original data. The feature number was set from 1 to 20 respectively to evaluate the performance of wavelet features and the original data. For Student's t-test, the best performance of 73.3% and 71.9% were obtained for 10 fold and 5 fold cross validation experiments running 50 times, when 6 features were selected from the original data. 11 wavelet features of CWT at scale 3, achieved 78.4% and 77.3% performance for 10 fold and 5 fold cross validation experiments respectively. Experimental results show that the performance of Student's t-test is better than one of Wilcoxon test. Table 2 shows the performance of the selected features based on Student's t-test and Wilcoxon test for CWT features and the original data.

From the above experiments, it is clear that CWT features obtain better performance than the original data. Because in wavelet space, the discriminant features between the two groups of patients are more significant than one in the original data space. Wavelet features reveal the properties which can not be detected in the original data space. Discrete wavelet transform extracts features and reduces the dimensionality of data. DWT extracts features at discrete position-scales ($2^1, 2^2, 2^3, 2^4, 2^5, \ldots$), not at every scale and position like CWT. The coefficients of CWT do not reduce dimensionality, but detect the features at every scale and act as complementary part of DWT. So CWT features achieve better performance than the original data or DWT features.

Table 2. The performance of CWT features and the original data based on Student's t-test and Wilcoxon test.

| method | t-test | | | Wilcoxon test | | |
|---|---|---|---|---|---|---|
| | FN | AC (10fold) | AC (5fold) | FN | AC (10fold) | AC (5fold) |
| Original data | 6 | 0.733 | 0.719 | 6 | 0.714 | 0.702 |
| CWT features | 11 | 0.784 | 0.773 | 5 | 0.738 | 0.738 |

FN represents feature number; AC represents accuracy.

7.3 The performance of the optimized GA features and the protein markers

From above experiments, it is clear that the CWT coefficients at scale 3 achieve the best 78.5% performance. In order to select the significant biomarkers, the optimized features are selected from wavelet coefficients of CWT at scale 3 based on genetic algorithm. Because CWT does not reduce dimensionality of data, and extracted features have the same dimensions as the original data, selected protein biomarkers can be located using the same index as the optimized CWT features. The performance of selected GA features and their corresponding protein markers are shown in Table 3. After removing the markers which have the same meaning, 18 protein markers are obtained to evaluate the performance based on Kaplan-Meier curve and Cox regression model.

Table 3. The performance of GA features and the corresponding protein markers.

| FN | Features (GA) | Accuracy (5 fold) |
|---|---|---|
| 5 | 'nuclear stat1', 'ulbp2 and 3', 'CD46', 'MHC2', 'caspase 3' | 0.779 |
| 6 | 'nuclear stat1', 'mica', ' raetig and ulbp1', 'CD46', 'caspase 3', 'MHC2' | 0.786 |
| 6 | 'vegfc', "caspase 3', ' raetie and ulbp1', 'CD46', 'ulbp2 and 3', 'MHC2' | 0.786 |
| 5 | 'CD46', 'ulbp1', 'nuclear stat1', 'RAETIG and RAETIE', 'mica and ulbp1' | 0.761 |
| 5 | 'CD46', 'ulbp1', 'nuclear stat1', 'caspase 3', 'mica and ulbp2' | 0.776 |
| 6 | 'b2m', 'CD46', 'MHC2', 'nuclear stat1', 'caspase 3', 'caspase3and p53' | 0.771 |
| 5 | 'CD46', 'MHC2', 'trail', 'caspase 3', 'mica and ulbp3' | 0.778 |
| 5 | 'CD46', 'MHC2', 'trailR2', 'caspase 3', 'raetie and ulbp1' | 0.767 |
| 6 | 'CD46', 'chk1', 'raetie and mica', 'nuclear stat1', 'caspase 3', 'raetie and ulbp1' | 0.779 |
| 6 | 'CD46', 'msi', 'ulbp1', 'nuclear stat1', 'caspase 3', 'tumour IL17' | 0.768 |
| 6 | 'CD46', 'chk1', 'nuclear stat1', 'caspase 3", 'p53 and trailR2', 'raetie and ulbp1' | 0.779 |
| 7 | 'nuclear stat1', 'CD46', 'CD68', 'ki67', 'FLIP', 'caspase 3', 'ulbp1' | 0.766 |

## 7.4 Survival analysis based on censored data

### 7.4.1 Using censored data

KM curve and Cox regression model were used to evaluate the performance of 18 selected protein markers. After removing the missing values of 18 protein markers, 246 patient samples were obtained. Among them, 100 patients died from colorectal cancer, and the other 146 patients were dead from post operative complications, or from non cancer related events, or are still alive. Figure 8(a) shows the histogram of patients dead from colorectal cancer. Figure 8(b) shows the other 146 patients. The patients in Figure 8(b) were separated into alive patients as shown in Figure 8(c), or the patients who died from unrelated causes as shown in Figure 8(d). 56 patient samples were removed including samples from the patients who died from post operative complications, or who died of other causes. There are 90 alive patients left in Figure 8(c), comprising 47% of the whole data. From Figure 8(a) and 8(c), it was observed that the patients whose survival time was between 40 and 70 months exceeded the number of dead patients from colorectal cancer during this period by a large margin. Some patients may die later, and we assumed that this number at $t_i$ is $\Delta x_i$. From formula (8), survival function $S$ is the product of the probabilities $p_i$ at $t_i$ .

where $p_i = 1 - \dfrac{d_i}{n_i}$. After considering the effect of the died number $\Delta x_i$ at $t_i$, the probabilities $p_i'$ at $t_i$ is calculated as:

$$p_i' = 1 - \frac{d_i + \Delta x_i}{n_i - \Delta x_i},$$

$$b_i = p_i' - p_i,$$

When $\Delta x_i$ is getting larger, the bias $b_i$ of probabilities at $t_i$ is getting higher. We only keep the patient samples surviving longer than 70 months and the censored data reduces to 32%, because most patients will survive after 70 months based on the distribution of dead patients in Figure 8(a).

Figure 9 shows KM curves of protein marker CD46 based on the different proportion of censored data. When the 32% censored data of alive patients with survival time of more than 70 months was used, CD46 was found to significant for survival (Figure 9(a)). This was consistent with the results of our proposed method of biomarker selection. When the 47% censored data of alive patients with survival time of more than 40 months were input into KM estimator, the results show that CD46 was not significant for survival (Figure 9(b)). This is because the number of censored data exceeds the number of complete data (dead patients) during survival time between 40 and 70 months by a larger margin, and this causes bias $b_i$.

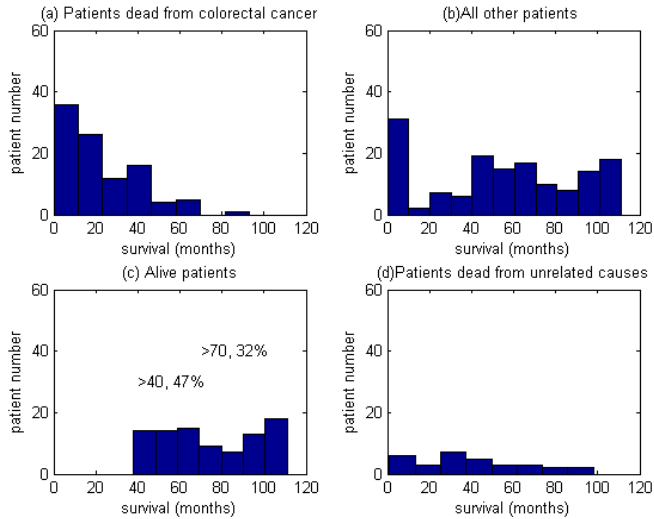

Fig. 8. The distribution of patients. (a) patients dead from colorectal cancer (b) other patients including patients dead from operation and other reasons, or patients are still alive. (c) alive patients. (d) patients dead from unrelated causes.

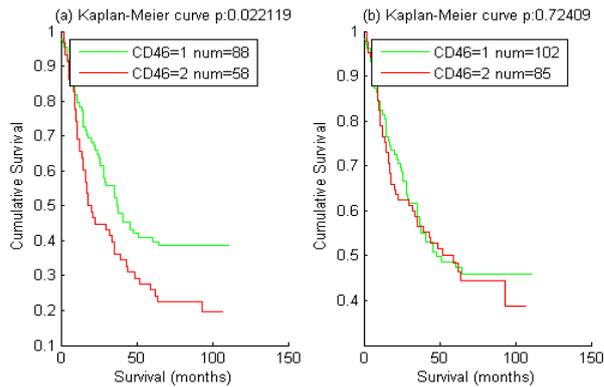

Fig. 9. Kaplan-Meier curves for disease-specific survival of CD46. (a) using the 32% censored data of alive patients with survival time of more than 70 months. (b) using the 47% censored data of alive patients with survival time of more than 40 months.

### 7.4.2 Survival analysis using Kaplan-Meier curve and multivariate Cox regression model

Kaplan-Meier curve with log-rank test was performed to evaluate the performance of each protein marker. The results are shown in Table 4. The protein markers CD46, chk1, p53 and trailR2, FLIP-L, nuclear stat1, and raet-IG and raet-IE show a significant difference in survival time. Some markers such as nuclear stat1, FLIP-L, p53, and mhc, have been found as prognostic biomarkers in colorectal cancer [1,2,45]. In [46], it is suggested that inhibition of chk1 kills tetraploid tumor cells through a p53-dependent pathway. Protein marker low expression of CD46 showed a mean survival time of 55.3 months (95% CI 47.2 to 63.4) which was significantly different to the mean survival time of 38.3 months (95% CI 29.5 to 47.1) for patients whose tumors express with high levels of CD46.

In order to evaluate the relative influence of CD46 expression with other biomarkers and with TNM stage, a multivariate analysis was performed using Cox regression model. The results were shown in Table 5. Protein marker CD46 shows independent prognostic significance with a hazard ratio of 1.787 (95% CI 1.126 to 2.836) for CD46 expression.

Based on recent research of Professor Lindy Durrant's group, it is suggested that the immune system might be involved in the development and progression of colorectal cancer [1, 14]. Professor Lindy Durrant explains that her team were "interested to see if the immune system, and in particular T cells and the IFN (interferon) γ pathway, was operational and influenced survival". Within next 2–3 years, the investigators are "planning to trial a new vaccine for colorectal cancer in patients with an intact immune system (60% of patients with colorectal cancer)". The detection of CD46 supports their conclusions based on recent research that CD46 is involving in the immune response [47, 48, 49, 50, 51, 52].

In [47], the paper focuses on current understanding of CD46 signaling in T-cell polarity and how this might influence disease outcome. It is indicated that the study of CD46 signaling in T cells has emerged as an exciting area of research that is shedding new light on how pathogens might manipulate the host immune response. A number of studies have shown that many tumors express higher levels of CD46. The expression of CD46 in primary cervical tissue shows a progressive increase from normal to malignant cells. An increased level of CD46 expression compared to normal tissue is also found in breast, endometrial, lung and hepatoma primary tumors [48]. As CD46 expression is up-regulated on numerous human cancers, viruses that bind CD46, such as measles virus and adenovirus, have been intensively investigated for their oncolytic potential in cancer therapy [47]. In [49], the human cell-surface molecule, CD46, has evolved from 'just another complement regulator' to a receptor for a striking array of pathogens. CD46 not only protects cells from complement-mediated attack and facilitates infection by a large number of pathogens, but also exerts complex effects on cellular immune function. In [50], the paper discusses the current knowledge about CD46 and its expanding roles in the immune system. In the last 10 years, CD46 is involved in a new and somewhat surprising functional aspect of the complement system: the down-modulation of adaptive T helper type 1 (Th1) immune responses by regulating the production of interferon (IFN)-$\gamma$ versus interleukin (IL)-10 within these cells. In [51], the authors summarize the latest updates on the regulation of CD46 expression and on its effects on T-cell activation. It is indicated that the last decade has revealed the role of CD46 in regulating the adaptive immune response, acting as an additional costimulatory molecule for human T cells and inducing their differentiation into Tr1 cells, a subset of regulatory T cells. In [52], the authors indicate that binding to CD46 can directly alter immune function and ligation of CD46 by antibodies or by measles virus can prevent activation of T cells by altering T-cell polarity and consequently preventing the formation of an immunological synapse. A mechanism by which CD46 reorients T-cell polarity to prevent T-cell receptor signaling in response to antigen presentation is defined.

Table 4. The performance of Kaplan-Meier curves for selected markers

| Variable | Category | Num | Mean survival month | 95% CI (months) | | P value |
|---|---|---|---|---|---|---|
| | | | | lower | upper | |
| 'CD46' | 1 | 88 | 55.3 | 47.2 | 63.4 | |
| | 2 | 58 | 38.3 | 29.5 | 47.1 | 0.022 |
| 'MICA ' | 1 | 69 | 41.0 | 32.4 | 49.5 | |
| | 2 | 77 | 55.5 | 46.9 | 64.0 | 0.083 |
| 'chk1' | 1 | 106 | 43.8 | 37.0 | 50.6 | |
| | 2 | 40 | 61.9 | 49.6 | 74.2 | 0.028 |
| 'b2m' | 1 | 26 | 44.1 | 28.9 | 59.3 | |
| | 2 | 120 | 49.8 | 43.1 | 56.4 | 0.702 |
| 'mhc' | 1,2 | 43 | 39.8 | 28.7 | 50.9 | |
| | 3 | 103 | 52.6 | 45.3 | 59.8 | 0.142 |
| 'p53and trailr2' | 0,1,2 | 76 | 38.6 | 31.2 | 45.9 | |
| | 3 | 70 | 59.8 | 50.4 | 69.2 | 0.003 |

| 'MLH1MSH2' | 0,1,2 | 48 | 38.7 | 28.7 | 48.6 | |
| | 3 | 98 | 53.8 | 46.3 | 61.3 | 0.066 |
| 'FLIP-L' | 1 | 139 | 50.7 | 44.4 | 57.0 | |
| | 2 | 7 | 14.4 | 2.6 | 26.3 | 0.027 |
| Nuclear Stat1 | 0 | 124 | 45.5 | 39.0 | 52.0 | |
| | 1,2,3 | 22 | 67.8 | 53.3 | 82.2 | 0.046 |
| 'MHC-II in stromal cells' | 0 | 82 | 45.9 | 37.8 | 54.1 | |
| | 1,2,3 | 64 | 52.2 | 43.1 | 61.3 | 0.370 |
| 'raetie and mica' | 0,1 | 43 | 42.5 | 31.4 | 53.7 | |
| | 2,3 | 103 | 51.2 | 44.0 | 58.5 | 0.328 |
| 'RAET-IG and RAET-IE' | 0,1 | 113 | 44.7 | 37.9 | 51.6 | |
| | 2,3 | 33 | 62.9 | 50.6 | 75.2 | 0.038 |
| 'ulbp2 and ulbp3' | 0 | 47 | 46.2 | 34.9 | 57.5 | |
| | 1,2 | 99 | 49.8 | 42.6 | 57.0 | 0.766 |
| 'mica and ulbp2' | 0,1 | 82 | 49.4 | 40.8 | 58.0 | |
| | 2 | 64 | 45.3 | 37.0 | 53.7 | 0.795 |
| 'RAET-IG and ulbp3' | 0 | 78 | 43.2 | 34.8 | 51.5 | |
| | 1,2 | 68 | 55.2 | 46.5 | 64.0 | 0.090 |
| 'caspase 3' | 0 | 81 | 50.2 | 42.0 | 58.3 | |
| | 1 | 65 | 47.1 | 38.0 | 56.3 | 0.728 |
| 'vegfc' | 0 | 103 | 48.3 | 41.0 | 55.6 | |
| | 1 | 43 | 50.2 | 39.2 | 61.2 | 0.718 |
| 'IL17 in tumours ' | 0 | 94 | 50.9 | 42.9 | 58.9 | |
| | 1,2,3 | 52 | 45.2 | 36.2 | 54.3 | 0.568 |

The biomarkers in red colour are found to be significant for survival based on p value (<0.05).

Table 5. The performance of Multivariate Cox regression model

| Variable | Category | Hazard ratio | P value | 95.0%CI for HR | |
| --- | --- | --- | --- | --- | --- |
| TNM | I and II III and IV | 2.912 | 0.000 | 2.175 | 3.900 |
| 'CD46' | 1 2 | 1.787 | 0.014 | 1.126 | 2.836 |
| 'MICA ' | 1 2 | 0.761 | 0.224 | 0.490 | 1.182 |
| 'chk1' | 1 2 | 0.811 | 0.452 | 0.470 | 1.399 |
| 'b2m' | 1 2 | 0.988 | 0.980 | 0.406 | 2.404 |
| 'mhc' | 1,2 3 | 0.846 | 0.395 | 0.576 | 1.243 |
| 'p53and trailr2' | 0,1,2 3 | 0.869 | 0.231 | 0.690 | 1.094 |
| 'MLH1MSH2' | 0,1,2 | 0.875 | 0.205 | 0.712 | 1.076 |
| | 3 | | | | |
| 'FLIP-L | 1 | 2.348 | 0.060 | 0.965 | 5.712 |
| | 2 | | | | |
| 'Nuclear Stat1' | 0 | 0.713 | 0.021 | 0.534 | 0.951 |
| | 1,2,3 | | | | |
| 'MHC-II in stromal cells' | 0 | 1.012 | 0.916 | 0.815 | 1.255 |
| | 1,2,3 | | | | |
| 'raetie and mica' | 0,1 | 1.032 | 0.855 | 0.735 | 1.449 |
| | 2,3 | | | | |
| 'RAET-IG and RAET-IE' | 0,1 | 0.946 | 0.832 | 0.564 | 1.586 |
| | 2,3 | | | | |
| 'ulbp2 and ulbp3' | 0 | 1.243 | 0.515 | 0.646 | 2.392 |
| | 1,2 | | | | |
| 'mica and ulbp2' | 0,1 | 0.835 | 0.546 | 0.464 | 1.501 |
| | 2 | | | | |
| 'RAET-IG and ulbp3' | 0 | 0.852 | 0.694 | 0.385 | 1.889 |
| | 1,2 | | | | |
| 'caspase 3' | 0 | 1.285 | 0.351 | 0.759 | 2.174 |
| | 1 | | | | |
| 'vegfc' | 0 | 1.091 | 0.564 | 0.812 | 1.465 |
| | 1 | | | | |
| 'IL17 in tumour cells' | 0 | 1.145 | 0.525 | 0.754 | 1.737 |
| | 1,2,3 | | | | |

This table shows the performance of Cox regression model for TNM stage and selected protein markers. The protein markers in red colour are independent prognostic biomarkers based on p value (<0.05).

## 8. Discussion and Conclusions

In this study we propose a novel method of biomarker detection in survival analysis. Two groups of patients were used to select the biomarkers of colorectal cancer data. One was the patients with survival time of less than 30 months, and another was the patients with survival time of more than 70 months. First continuous wavelet analysis was used to extract the discriminant features between the two groups of patients. The best discriminant features were obtained based on CWT at scale 3. Genetic algorithm was performed on extracted wavelet coefficients to select the optimized features, using Bayes classifier in its fitness function. The best performance of 78.6% was obtained based on 6 optimized wavelet features, using 5 fold cross validation experiments. After genetic algorithm was runing several times to select the best features, several groups of biomarkers were detected. Some of the data referred to the same biomarker was removed, and 18 unique biomarkers were selected. Kaplan-Meier curve and Cox regression models were used to evaluate the performance of selected biomarkers. Protein markers CD46, chk1, p53, FLIP-L, nuclear stat1, RAET-IG and RAET-IE were found to be significant for survival using KM estimator with log-rank test. Cox regression model showed that CD46 and nuclear stat1 were independent prognostic biomarkers.

The proportion of censored data affects the selection of biomarkers in survival analysis. Protein marker CD46 was found to take an important role in survival analysis for colorectal cancer patients, using machine learning methods of wavelet analysis and genetic algorithm. KM curve shows that CD46 was not significant for survival using 47% censored data with survival time above 40 months, but was significant using 32% censored data for survival with survival time of more than 70 months. In this study the number of censored data with survival time ranging from 40 to 70 months was much more than one of uncensored data (dead patients). This causes a bias affecting KM curve and thus some significant biomarkers were not detected. Machine learning methods reveal the hidden information of colorectal cancer data, which cannot be detected by traditional survival analysis methods, in particular using KM curve and Cox regression model with a large proportion of censored data.

One of the primary innate mechanisms to prevent tumor growth is activation of the complement cascade. Activation of complement occurs via a cascade of enzyme activity, initiated by either the antibody-dependant classical pathway, or the antibody-independent alternative and lectin pathways [53]. These lead to a common activation of the C3 component of complement, and in turn to the formation and membrane insertion of a terminal C5b-9 membrane attack complex (MAC), causing direct lysis of the target cell. To protect themselves from bystander attack by complement cells express membrane-bound complement regulatory proteins (mCRP) which act predominately at either the C3/C5 convertase level as with membrane cofactor protein (MCP; CD46) and decay accelerating factor (DAF; CD55), or act further downstream to inhibit assembly of the MAC as with protectin (CD59). Expression of one or more mCRP (frequently at a greater level than the corresponding normal tissue) has been demonstrated for most solid tumor types [54] and confers resistance to tumor elimination by complement dependent mechanisms.

The mCRP CD46 has been identified on all human cells exposed to complement except erythrocytes [55]. Unlike CD55 and CD59 which are GPI-anchored, the CD46 molecule inserts into the cell membrane via a transmembrane domain [56], acting as a cofactor for the factor-I-mediated cleavage of C3b and C4b into inactive forms and clearing these molecules from the surface of host cells. Previous attempts to characterise CD46 expression in both normal and neoplastic tissues and in tumor cell lines have been limited to the analysis of small numbers of cases. In the case of colorectal tissues, expression of CD46 by normal colonic epithelium and colonic adenomas has been shown to display a predominately basal and baso-lateral membrane staining, with circumferential membrane CD46 expression seen in colonic adenocarcinomas [57]. Both Koretz et al. [57] and Thorsteinsson et al. [58] described consistent membrane expression of CD46 in analyses of 71 and 18 colonic adenocarcinomas respectively, with both sets of authors noting a higher antigen density in the neoplastic compared with non-neoplastic epithelium leading to the conclusion that CD46 is generally upregulated during malignant colorectal tumour progression. Similar findings of ubiquitous CD46 expression have been reported for tumors of the breast [59] and stomach [60].

Experimental results show that our proposed method, which combines machine learning methods of wavelet analysis, genetic algorithm, and Bayes classifier with survival analysis methods of KM curve and Cox regression model, provides an efficient way to select potentially significant prognosis markers. A new protein marker CD46 was found significant in survival based on our proposed method.